%% file: main.tex
\documentclass[times, review, 10pt]{elsarticle}
\usepackage{graphicx,verbatim}
\usepackage{amsmath}
\usepackage{algorithmic}
\usepackage{algorithm}
\usepackage{multirow}
\usepackage{amssymb}
\usepackage{xcolor}
\usepackage{colortbl}
\usepackage{array}
\usepackage{booktabs}
\usepackage{hyperref}
\hypersetup{
    colorlinks=true,
    linkcolor=blue,
    citecolor=blue,
}

\newcommand{\ie}{\emph{i.e}.}
\newcommand{\eg}{\emph{e.g}.}

\input{math_commands.tex}

\journal{Pattern Recognition}

\begin{document}

\begin{frontmatter}

\title{BeatDance: Generating Beat-Consistent 3D Dance with Hierarchical Spatial-Temporal Modeling}

\author[label1]{Xiaojian Shen}
\author[label2]{Dahu Shi}
\author[label3]{Jianrong Zhang}
\author[label4]{Hai Li}
\author[label4]{Hongwei Zhao}
\author[label5]{Dawei Zhang}
\author[label6]{Yunzhi Zhuge}
\author[label2]{\corref{cor1}Zhiliang Wu}
\author[label9]{Guanghui Yue}
\author[label7]{Wei Zhou}
\cortext[cor1]{Corresponding author: Zhiliang Wu}

\affiliation[label1]{organization={College of Software, Jilin University},
            city={Changchun},
            postcode={130012},
            country={China}}

\affiliation[label2]{organization={College of Computer Science and Technology, Zhejiang University},
            city={Hangzhou},
            postcode={310007},
            country={China}}

\affiliation[label3]{organization={ReLER, AAII, University of Technology Sydney},
            city={Sydney},
            country={Australia}}

\affiliation[label4]{organization={College of Computer Science and Technology, Jilin University},
            city={Changchun},
            postcode={130012},
            country={China}}

\affiliation[label5]{organization={School of Computer Science and Technology, Zhejiang Normal University},
            city={Jinhua},
            postcode={321004},
            country={China}}

\affiliation[label6]{organization={School of Information and Communication Engineering, Dalian University of Technology},
            city={Dalian},
            postcode={116024},
            country={China}}
\affiliation[label9]{organization={School of Biomedical Engineering, Shenzhen University Medical School, Shenzhen University},
            city={Shenzhen},
            postcode={518060},
            country={China}}
\affiliation[label7]{organization={School of Computer Science and Informatics, Cardiff University},
            city={CF10 3AT Cardiff},
            country={U.K.}}

\begin{abstract}
Generating realistic 3D dance from music is a challenging task that requires accurate synchronization with musical rhythms while capturing the spatial complexity of human motion. Although existing methods can generate physically plausible dance motions, they often struggle to achieve precise alignment with music, such as the beat. To address this limitation, we propose a novel diffusion-based framework, BeatDance, with two components: 1) We present a Hierarchical Decoupled Attention (HDA) module, which first disentangles the learning of human pose and temporal dynamics. A hierarchical structure is then employed to capture both short-term and long-term dependencies, thereby enhancing spatial-temporal modeling. 2) We adopt cycle-consistent learning by introducing an auxiliary dance-to-music module. During training, discrepancies between the reconstructed and original music induce a stronger loss signal, effectively encouraging the consistency property between the music and dance motion.
Extensive experimental results demonstrate that our proposed approach outperforms recent competitive methods on two benchmark datasets.
The project page is available at \url{https://shenxiaojian.github.io/BeatDance/}.
\end{abstract}


\end{frontmatter}

\section{Introduction}
Music plays an important role in shaping emotions and narratives, which are often embodied through human motion. This natural connection has inspired growing interest in generating 3D dance conditioned on the music. Automated motion synthesis from music reduces reliance on motion-capture systems and costly animation tools, enabling more efficient content creation across various applications such as film-making~\cite{gu2024learning,wu2025bvinet}, virtual production~\cite{jia2026gas}, and video editing~\cite{wu2024waveformer,wang2024text}.

\begin{figure}[!b]
\centering
\includegraphics[width=0.6\columnwidth]{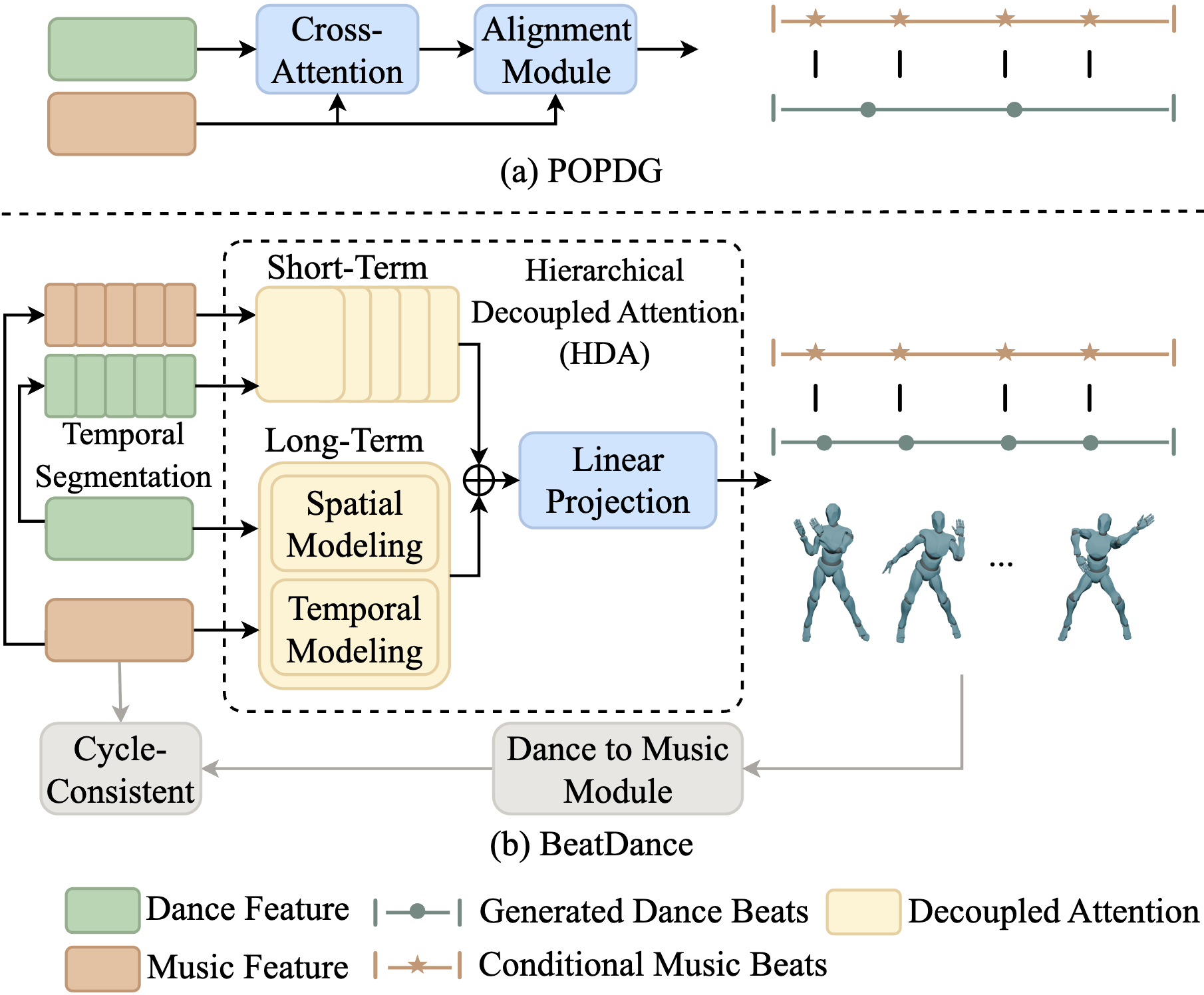}
\caption{Comparison with the existing state-of-the-art method. (a) POPDG~\cite{luo2024popdg} uses a cross-attention and an alignment module to associate music and dance. However, it may face challenges in achieving beat-level synchronization. (b) We propose Hierarchical Decoupled Attention (HDA) and a cycle-consistent learning strategy. HDA hierarchically decouples spatial and temporal information, while cycle-consistent learning leverages the inverse dance-to-music mapping to further enhance temporal alignment.}
\label{intro}
\end{figure}

To achieve high-quality dance generation, many works leverage the autoencoder~\cite{siyaoduolando,xu2024gg} as a core component of the model architecture. Bailando~\cite{siyao2022bailando} models the upper and lower body motions separately using VQ-VAE, and generates motion code indices auto-regressively. TM2D~\cite{gong2023tm2d} combines the training of music-to-dance and text-to-motion, which helps produce smoother dance motion.
{Recently, diffusion models have shown remarkable proficiency in music-driven 3D dance generation~\cite{zhang2024bidirectional,li2024lodge,li2024lodge++,wang2024explore}.
For example, EDGE~\cite{tseng2023edge} proposes to use Jukebox~\cite{dhariwal2020jukebox} to extract music features, and the model also supports dance editing, dance completion, and joint in-painting.
POPDG~\cite{luo2024popdg} proposes a new dataset, \ie, PopDanceSet, which comprises a wide range of diverse and complex dance motions. It introduces an additional self-attention and an alignment module to learn the relationships among human body parts and insert music features into the model, respectively.}

{
Despite the encouraging progress, these methods still face challenges in generation precision.
First, they only perform a coarse-grained fusion of music and motion features without explicitly modeling the spatial-temporal dependencies simultaneously between music and dance, which lead to beat misalignment {(such as POPDG illustrated in Figure~\ref{intro}(a))}. Second, these works typically focus on long-term dependency, overlooking the short-term patterns that are crucial for maintaining motion continuity and synchronizing with local-level musical beats.}

To address these limitations, we propose BeatDance, a diffusion-based framework for music-driven 3D dance generation.
At the core of this framework is our proposed Hierarchical Decoupled Attention (HDA), which consists of two components.
First, we present a decoupled attention that explicitly separates the learning of spatial and temporal information. Specifically, the spatial attention divides the body into several anatomical parts, each with an independent linear projection. For the temporal dynamics, we update the temporal feature by applying a Taylor series expansion over a learned joint music-dance attention map. Second, we employ a hierarchical design that organizes these motion features from local to global levels for short-term and long-term dependencies. This design provides the capacity to model fine-grained music-dance correspondences across multiple temporal scales.
Meanwhile, to further exploit this capacity, we introduce a dance-to-music branch based on a latent diffusion model. We perform cycle-consistent learning to map dance motion back to its corresponding music, which improves the alignment between the two modalities. The improved music-dance alignment is particularly reflected in more accurate beat synchronization. It also benefits broader aspects of dance quality: the part-specific spatial modeling facilitates the learning of complex dance poses and expressive whole-body movements, while the short-term and long-term temporal modeling promote smooth transitions and choreographic coherence, respectively.

Taking these components together, our approach is able to generate expressive, smooth, and choreographically coherent dance motions that are well synchronized with the music, particularly aligning closely with the underlying beat (as shown in Figure~\ref{intro}(b)). We conduct extensive experiments on two datasets, PopDanceSet and AIST++. For example, on the PopDanceSet dataset, BeatDance achieves a PBC of 7.91 and BAS of 0.244, outperforming POPDG of 5.95 and 0.233, respectively.

In summary, our contributions include:
\begin{itemize}
    \item We present BeatDance, a diffusion-based framework for music-driven 3D dance generation that achieves state-of-the-art physical plausibility and beat alignment on both benchmarks, while maintaining competitive motion diversity.
    \item We propose a Hierarchical Decoupled Attention (HDA), which separately models motion spatial structures and temporal dynamics. It also incorporates a hierarchical architecture to capture both short-term and long-term dependencies between the music and dance.
    \item We introduce a cycle-consistent learning framework by predicting music features from generated dance motion through a dance-to-music module, which further enhances alignment with the musical beat.
\end{itemize}

\section{Related Works}
In this section, we review works closely related to ours in two areas. We first discuss text-driven motion generation, which shares similar architectural foundations with our approach. We then review music-driven dance generation, with a focus on recent diffusion-based methods and their limitations in achieving precise cross-modal alignment.
\subsection{Text-Driven Motion Generation}
Text-driven motion generation aims to synthesize realistic human motion from a text description.
Early works~\cite{petrovich22temos,chuan2022tm2t}
directly deploy a text encoder and a motion decoder to learn the mapping from the two modalities.
Recently, some works~\cite{lu2024humantomato,pinyoanuntapong2024mmm} employ a Vector Quantized Variational Autoencoder (VQ-VAE)~\cite{oord2017neural} to project human motions into discrete representations.
For example, T2M-GPT~\cite{zhang2023t2m} encodes motion sequences into discrete tokens using VQ-VAE and then learns to map text descriptions to these tokens with a Generative Pretrained Transformer (GPT).
Another trend is the use of diffusion models for motion synthesis. MotionDiffuse~\cite{zhang2024motiondiffuse} and MDM~\cite{tevet2023human} are among the first efforts to apply diffusion models~\cite{ho2020denoising} to this field.
Subsequent works achieve performance improvements by encoding the motion into latent space~\cite{zhang2024motion,yue2025text,khamis2025i2fet}, adopting a retrieval-augmented generation strategy~\cite{zhang2023remodiffuse}, and exploring spatial-temporal decoupled attention~\cite{chen2026smrnet,yue2023attention,ma2026uncertainty,wu2026dlvinet,yue2023dual,wu2023deep,ding2024ksof}.
For example, MLD~\cite{chen2023executing} learns low-dimensional latent vectors of a motion sequence with a Variational Autoencoder (VAE). Then, it employs a latent diffusion model to generate latent vectors from text. FineMoGen~\cite{zhang2023finemogen} proposes a spatio-temporal mixture attention to separately model spatial and temporal dependencies. Although it shares a related modeling principle with our method, our work focuses on the distinct challenge of fine-grained music-dance alignment: the generated dance should respond to rapidly changing local rhythmic patterns while maintaining long-term motion coherence. To address this challenge, our HDA explicitly models fine-grained relationships between the two modalities and hierarchically captures short-term rhythmic patterns and long-term dependencies.
Furthermore, we propose a cycle-consistent training paradigm that further improves the music-dance alignment.

\subsection{Music-Driven Dance Generation}

Music-driven dance generation is challenging due to its demand for complex articulation and precise music-dance alignment.
Early methods~\cite{lee2013music, ofli2011learn2dance}, which rely on similarity-based retrieval and concatenation from existing databases, are inherently limited in generating novel dance motions.
With the advent of deep learning, extensive efforts~\cite{wu2021music-to-dance,huang2022genre-conditioned,li2021autodance,lee2022human} have been made to model the probabilistic mapping between music and dance.
Among these advances, VQ-VAEs have emerged as an effective tool for learning discrete motion representations.
For example, Bailando~\cite{siyao2022bailando,siyao2023bailandopp} encodes motion into discrete codes and uses a GPT-style autoregressive decoder to generate upper- and lower-body movements, with a reinforcement-learning-based scheme for beat alignment.
Similarly, TM2D~\cite{gong2023tm2d} utilizes a shared VQ-VAE codebook for both music-to-dance and text-to-motion tasks. By unifying the shared latent space, it benefits from large-scale text-motion datasets, resulting in smoother dance motions.
More recently, diffusion models have shown remarkable proficiency in music-driven 3D dance generation~\cite{zhang2024bidirectional,wang2024explore,zhang2025danceeditor}, branching into two distinct research directions.
One direction focuses on extremely long dance generation via multi-stage architectures~\cite{li2024lodge,li2024lodge++}. For example, Lodge~\cite{li2024lodge} employs a coarse-to-fine framework to achieve this through cascaded generation processes. Different from these multi-stage approaches, our HDA captures short-term and long-term dependencies within a unified single-stage module, with a particular focus on improving fine-grained music-dance alignment.
The other direction explores single-stage frameworks to streamline the generation pipeline.
EDGE~\cite{tseng2023edge} is the first to leverage the diffusion model for synthesizing high-fidelity dance in this category.
It uses Jukebox~\cite{dhariwal2020jukebox} to extract music features, which has become the standard setting for subsequent works.
POPDG~\cite{luo2024popdg} proposes PopDanceSet, a dataset of diverse and complex dance motions. It also introduces spatial-temporal attention blocks and an alignment module that modulates dance features through an affine transformation conditioned on music and motion features. By contrast, our HDA explicitly models fine-grained music-dance correspondences. It first partitions the body into anatomical regions with independent projections, allowing different body parts to separately capture their relationships with the music. It then constructs a joint music-dance attention map and leverages a higher-order Taylor expansion to model complex temporal dynamics. Furthermore, the hierarchical design captures short-term rhythmic patterns and long-term dependencies, enabling the generated motion to respond to local musical beats while maintaining coherent whole-body dynamics. Therefore, while POPDG fuses music and dance features through an affine modulation of dance features, HDA explicitly models how the relationships between music and different body parts evolve across multiple temporal scales.

\section{Preliminaries}
\label{sec:preliminaries}

Denoising Diffusion Probabilistic Models (DDPM)~\cite{ho2020denoising} are a class of generative models that learn to synthesize data by reversing a gradual noising process.
In this section, we briefly review the key concepts of DDPM that form the foundation of our framework.

\paragraph{Forward process}
Given a data sample $\mX_0 \sim q(\mX_0)$, the forward process gradually corrupts it by injecting Gaussian noise over $T$ timesteps.
By defining the cumulative noise schedule $\alpha_t = \prod_{s=1}^{t}(1 - \beta_s)$, where $\{\beta_t\}_{t=1}^{T}$ is a fixed variance schedule, the noisy sample at an arbitrary timestep $t$ can be obtained in closed form:
\begin{equation}
    \mX_t = \sqrt{\alpha_t}\,\mX_0 + \sqrt{1 - \alpha_t}\,\epsilon,
    \quad \epsilon \sim \mathcal{N}(0, 1).
    \label{eq:forward}
\end{equation}
When $t = T$, $\alpha_T \approx 0$, so $\mX_T \sim \mathcal{N}(\mathbf{0}, \mathbf{I})$ is approximately pure Gaussian noise.

\paragraph{Reverse process}
The reverse process learns a denoising network $\epsilon_\theta$ to iteratively recover the original data from $\mX_T$.
Given a conditioning signal $\vc$ (\eg, music features in our framework), the network predicts the original data $\mX_0$ from the noisy input $\mX_t$ at each step.
The training objective is:
\begin{equation}
    \mathcal{L}_{\text{diff}}
    = \mathbb{E}_{\mX_0,\,\vc,\,\epsilon,\,t}\Big[
        \Vert \mX_0 - \epsilon_\theta(\mX_t, t, \vc) \Vert_{2}^{2}
      \Big].
    \label{eq:ddpm_loss}
\end{equation}
Following~\cite{nichol2021iddpm}, the reverse variance can also be learned jointly via a variational lower-bound loss $\mathcal{L}_\text{vlb}$ to better model the reverse transition distribution.
At inference time, we adopt DDIM~\cite{song2021ddim}, which reformulates the reverse process as a non-Markovian procedure and enables high-quality generation with significantly fewer denoising steps.

\section{Method}
In this section, we first introduce BeatDance, a diffusion-based framework designed to synthesize realistic and expressive 3D dance sequences that are consistent with the source music. Then we present Hierarchical Decoupled Attention (HDA), which is used to model the intricate spatial-temporal mappings between generated dance and input music, capturing both short-term and long-term dependencies in a hierarchical fashion. Finally, we propose Cycle-Consistent Learning (CCL) mechanism to reinforce the music-dance consistency by employing an auxiliary dance-to-music model.

\begin{figure*}[!t]
\centering
\includegraphics[width=\textwidth]{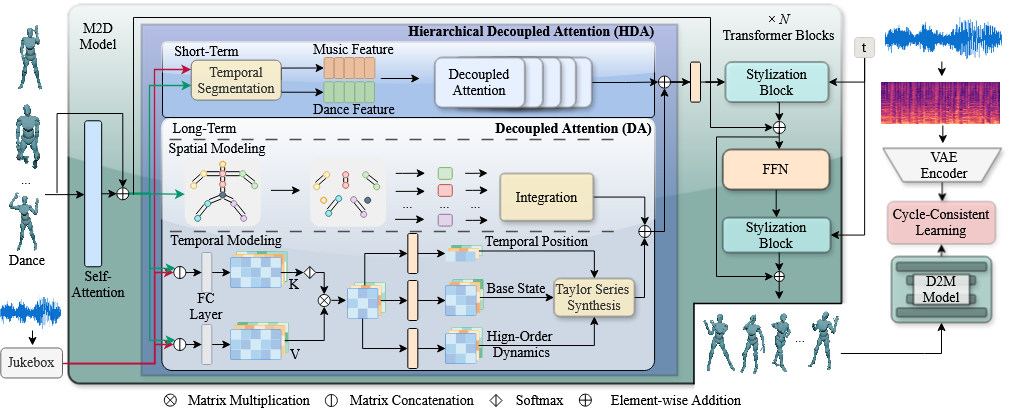}
\caption{Overview of BeatDance. Hierarchical Decoupled Attention (HDA) employs a hierarchical design to capture both short-term and long-term music-dance dependencies, building upon a Decoupled Attention (DA) that splits the modeling into spatial and temporal branches. Additionally, our cycle-consistent learning mechanism uses an auxiliary Dance-to-Music (D2M) model to reconstruct the source music features from the generated dance, thereby enforcing a tighter music-dance alignment.}
\label{overview}
\end{figure*}

\subsection{Overview}
\label{subsec:sec3_diffusion}
The overall architecture is shown in Figure~\ref{overview}, which consists of a Music-to-Dance (M2D) model and a Dance-to-Music (D2M) model. We adopt a skeleton-based diffusion architecture for the former, which is trained from scratch with music as the conditioning input. The latter is used to reconstruct the music from the generated dance.

{Specifically, in the music-to-dance branch, we follow the standard music-driven 3D dance generation setting to use Jukebox~\cite{dhariwal2020jukebox} to extract the music embedding from a piece of music. Given a music embedding $\mM = [\vm^1,\vm^2,\cdots,\vm^{T}]$ and a dance motion sequence $\mX = [\vx^1,\vx^2,\cdots,\vx^{T}]$, where $\vm^i \in \mathbb{R}^{1 \times d_m}$, $\vx^i \in \mathbb{R}^{1 \times d_x}$, $T$ is the length, which is common to both the music feature and the dance sequence. We employ a diffusion-based model to learn the mapping from the music embedding to the corresponding dance motion. In the forward process, noise $\epsilon \sim \mathcal{N}(0,1)$ is incrementally added to the dance sequence, obtaining $\mX_t = \sqrt{\alpha_t} \mX_0 + \sqrt{1-\alpha_t} \epsilon$ at timestep $t$. Then, we feed $\mX_t$ into a transformer-based denoising autoencoder $\epsilon_\theta^s$ to predict the original dance sequence conditioned on $\mM$. The model $\epsilon_\theta^s$ incorporates a Hierarchical Decoupled Attention (HDA) mechanism, which separately models spatial (pose-wise) and temporal dependencies in a hierarchical fashion. This mechanism facilitates the learning of fine-grained correlations between music and dance and enhances the beat alignment (Section~\ref{subsec:hda}). The optimization of $\epsilon_\theta^s$ is performed using $\mathcal{L}_{\text{m2d}}$, which can be written as
\begin{equation}
\mathcal{L}_{\text{m2d}} = \mathcal{L}_{\text{diff}} + \lambda_{\text{vlb}}\mathcal{L}_{\text{vlb}} +
\lambda_{\text{va}}\mathcal{L}_{\text{va}}  + \lambda_{\text{joint}}\mathcal{L}_{\text{joint}} + \lambda_{\text{body}}\mathcal{L}_{\text{body}},
\notag
\end{equation}
where $\mathcal{L}_{\text{diff}} = \mathbb{E}_{\mX, \mM, \epsilon, t}\Big[ \Vert \mX - \epsilon_\theta^s(\mX_t, t, \mM)\Vert_{2}^{2}\Big]$. We incorporate a variational lower-bound loss, denoted as $\mathcal{L}_{\text{vlb}}$, weighted by a coefficient $\lambda_{\text{vlb}}$, to learn the variance of the diffusion process transitions~\cite{nichol2021iddpm}. Following~\cite{tseng2023edge,luo2024popdg}, we also use three Mean Squared Error (MSE) losses, \ie, $\mathcal{L}_{\text{va}}$, $\mathcal{L}_{\text{joint}}$, and $\mathcal{L}_{\text{body}}$. These losses respectively constrain velocity and acceleration, encourage the consistency in joint space, and enhance ground contact for feet, hands, and the neck. $\lambda_{\text{va}}$, $\lambda_{\text{joint}}$ and $\lambda_{\text{body}}$ are hyperparameters that balance the weight of each loss.}

{As for the dance-to-music branch, we aim to reconstruct the music from the generated dance motion with a latent diffusion model. Building upon this model, we introduce a cycle-consistent learning strategy to further encourage semantic consistency between the generated 3D dance and the source music. Briefly, we first train the dance-to-music model independently. Once trained, we freeze its parameters and integrate it with the music-to-dance model to form a music-to-dance-to-music cycle. For more details, please refer to Section~\ref{dance-to-music}.
}

\textbf{Overall optimization goal.} {Taking these components together, our training objective can be formulated as follows:
\begin{equation}
\mathcal{L}_{\text{total}} = \mathcal{L}_{\text{m2d}} + \lambda_{\text{cycle}}\mathcal{L}_{\text{cycle}},
\label{formula:overall_loss}
\end{equation}
where $\mathcal{L}_{\text{cycle}}$ denotes the cycle-consistent loss (Section~\ref{cycle}), weighted by a coefficient $\lambda_{\text{cycle}}$.}

\subsection{Hierarchical Decoupled Attention}
\label{subsec:hda}
{Different from using cross-attention and alignment modules that jointly model spatial and temporal information, we propose a Hierarchical Decoupled Attention (HDA), which explicitly disentangles spatial and temporal modeling (Decoupled Attention) and captures both short-term and long-term dependencies (Hierarchical Mechanism).}

\subsubsection{Spatial Modeling}
To effectively learn complex dance poses, we employ an anatomical partitioning strategy.
We divide each frame's raw representation into $N_s=8$ parts, comprising 7 functionally relevant regions that incorporate the head, spine, left arm, right arm, left leg, right leg, root joint, and an additional global part representing the full body.
{Each part is fed into an independent linear projection layer ($N_s$ in total), and the outputs are concatenated to obtain $\mZ \in \mathbb{R}^{T \times N_s \times d_s}$, where $d_s$ represents the dimension of each human body part. Then, a Feed-Forward Network (FFN) is applied to each body part independently. To enable interaction across different parts, we employ a learnable parameter matrix $\mW^s \in \mathbb{R}^{N_s \times N_s}$ for the feature integration. Specifically, at the $t$-th time step, the spatial feature is computed as
\begin{equation}
    \mZ^\text{out} = \mW^s \cdot \text{FFN}(\mZ),
\end{equation}
where $\cdot$ denotes the matrix multiplication. Finally, the spatial feature is reshaped into $\mZ^{\text{spatial}} \in \mathbb{R}^{T \times d'_s}$ with $d'_s = N_s \times d_s$, which serves as the final output.}

\subsubsection{Temporal Modeling}
{We first map the dance feature $\mX$ and music embedding $\mM$ to keys ($\mK^d$, $\mK^m$) and values ($\mV^d$, $\mV^m$) via linear projection and concatenate them to obtain $\mK = [\mK^d; \mK^m] \in \mathbb{R}^{2T \times d_k}$ and $\mV = [\mV^d; \mV^m] \in \mathbb{R}^{2T \times d_v}$, where $d_k$ and $d_v$ are the dimensions of key and value, respectively. Note that $d_v = d'_s$. Then, the attention map $\mA \in \mathbb{R}^{d_k \times d_v}$ is computed as $\mA = \text{softmax}(\mK^\top)\mV$, which effectively encodes the music and dance information into $d_k$ distinct latent bins. Each bin is characterized by a $d_v$-dimensional feature vector, which serves as the basis for our temporal dynamics modeling.}

{Specifically, $\mA$ is processed by 5 distinct FFNs to produce a set of dynamic components: a temporal position matrix $\hat{\mT} \in \mathbb{R}^{d_k \times 1}$, a base state matrix $\mF_0 \in \mathbb{R}^{d_k \times d_v}$, and three derivative matrices $\mF_1, \mF_2, \mF_3 \in \mathbb{R}^{d_k \times d_v}$, which encode the scaled first-, second-, and third-order motion dynamics, respectively. Following~\cite{zhang2023finemogen}, we update the attention map using a Taylor expansion-based mechanism, which allows the model to incorporate higher-order temporal variations in a structured and interpretable manner:
\begin{equation}
\mZ^{\text{TS}} = \mF_0 + \mF_1 \odot (\hat{\vt} - \hat{\mT}) + \mF_2 \odot (\hat{\vt} - \hat{\mT})^2 + \mF_3 \odot (\hat{\vt} - \hat{\mT})^3,
\end{equation}
where $\odot$ denotes element-wise multiplication, and the $\hat{\vt}=[\hat{t}^1, \hat{t}^2, \cdots, \hat{t}^T] \in \mathbb{R}^T$ with $\hat{t}^i \in \mathbb{R}$ is the time sequence vector. Please find more details about the broadcasting process in Algorithm~\ref{alg:taylor_expansion}.}

{To integrate $d_k$ latent bins in the $\mZ^{\text{TS}} \in \mathbb{R}^{T \times d_k \times d_v}$ into a unified representation, we employ a dynamic weighting strategy. This approach assigns a relevance score to each bin based on its temporal proximity to the current frame, allowing the model to selectively emphasize temporally coherent features. Formally, for each timestep $t$, the temporal feature $\mZ^\text{temporal}_t \in \mathbb{R}^{1 \times d_v}$ is computed via a weighted sum over $d_k$ bins:
\begin{equation}
\label{eq:weighted_sum}
\mZ^{\text{temporal}}_{t} = \sum_{k=1}^{d_k} \gamma_{t,k} \cdot \mZ^{\text{TS}}_{t,k},
\end{equation}
where $\gamma_{t,k}$ and $\mZ^{\text{TS}}_{t,k}$ denote the $(t,k)$-th entry of the weight matrix $\mathbf{\Gamma} \in \mathbb{R}^{T \times d_k}$ and $\mZ^{\text{TS}}$, respectively. Then, $\gamma_{t,k}$ is processed by a softmax-normalized Gaussian kernel:
\begin{equation}
\label{eq:gamma_weights}
\gamma_{t,k} = \frac{\exp(-(t - \hat{\mT}_{k})^2 / \sigma^2)}{\sum_{i=1}^{d_k} \exp(-(t - \hat{\mT}_{i})^2 / \sigma^2)},
\end{equation}
where $\sum_{k=1}^{d_k} \gamma_{t,k} = 1$, $\hat{\mT}_{k}$ denotes the temporal position associated with the $k$-th bin, and $\sigma^2$ is a variance hyperparameter that controls the smoothness of the weighting distribution.
In this way, the model can focus more on features that are temporally aligned with the current context.}

{Finally, the full temporal representation $\mZ^{\text{temporal}} \in \mathbb{R}^{T \times d_v}$ is obtained by concatenating the frame-wise features $\mZ^{\text{temporal}}_t$ along the temporal dimension.}

\subsubsection{Spatio-Temporal Feature Fusion}
With $\mZ^\text{spatial}$ and $\mZ^\text{temporal}$ available, we integrate them via an additive operation,
\begin{equation}
\mZ^{\text{global}} = \mZ^\text{spatial} + \mZ^\text{temporal}.
\end{equation}
Such a fusion allows the model to jointly leverage spatial-temporal configuration, resulting in a more informative global-level representation.

\subsubsection{Hierarchical Mechanism}
We also propose a hierarchical mechanism to capture long-term and short-term music-dance dependencies. The aforementioned global feature $\mZ^{\text{global}}$ is used to represent long-term dependencies, while short-term dynamics are modeled within local-level temporal segments. Specifically, we first partition the music and dance features into $K$ non-overlapping short-term windows along the temporal dimension, obtaining $\hat{\mM} = [\mM_l^1, \mM_l^2,\cdots,\mM_l^K]$ and $\hat{\mX} = [\mX_l^1, \mX_l^2,\cdots,\mX_l^K]$. We independently apply the decoupled attention mechanism to each window. Subsequently, the features extracted from all windows are concatenated along the temporal axis to form the local-level representation $\mZ^{\text{local}}$.
Finally, to incorporate hierarchical information, global and local representations can be combined as follows:
\begin{equation}
\mZ^{\text{h}} = \text{FFN}(\mZ^{\text{global}} + \mZ^{\text{local}}),
\end{equation}
where $\mZ^{\text{h}}$ denotes the output of our Hierarchical Decoupled Attention (HDA) module, which effectively captures both global dynamics and local details.

\begin{algorithm}[htb]
\caption{Broadcasting Details in Taylor Expansion}
\label{alg:taylor_expansion}
\small
\textbf{Input}: \\
\hspace*{\algorithmicindent} Time sequence vector $\hat{\vt} \in \mathbb{R}^{T}$; \\
\hspace*{\algorithmicindent} Temporal position matrix $\hat{\mT} \in \mathbb{R}^{d_k \times 1}$; \\
\hspace*{\algorithmicindent} Base state matrix $\mF_0 \in \mathbb{R}^{d_k \times d_v}$; \\
\hspace*{\algorithmicindent} Dynamics matrices $\{\mF_i\}_{i=1}^3 \in \mathbb{R}^{d_k \times d_v}$.\\
\textbf{Output}: Final output matrix $\mZ^{\text{TS}} \in \mathbb{R}^{T \times d_k \times d_v}$. \\
\small
\begin{algorithmic}[1] \STATE \textit{// Reshape inputs to enable broadcasting}
\STATE $\hat{\vt}' \leftarrow \text{reshape}(\hat{\vt}, (T, 1, 1))$
\STATE $\hat{\mT}' \leftarrow \text{reshape}(\hat{\mT}, (1, d_k, 1))$
\STATE $\mF'_0 \leftarrow \text{reshape}(\mF_0, (1, d_k, d_v))$
\STATE $\mF'_1 \leftarrow \text{reshape}(\mF_1, (1, d_k, d_v))$
\STATE $\mF'_2 \leftarrow \text{reshape}(\mF_2, (1, d_k, d_v))$
\STATE $\mF'_3 \leftarrow \text{reshape}(\mF_3, (1, d_k, d_v))$

\STATE \textit{// Compute the time difference via broadcasting}
\STATE $\Delta \leftarrow \hat{\vt}' - \hat{\mT}'$ \COMMENT{Resulting shape is $(T, d_k, 1)$}

\STATE \textit{// Compute all terms and sum them up using broadcasting}
\STATE $\mZ^{\text{TS}} \leftarrow \mF'_0 + \mF'_1 \odot \Delta + \mF'_2 \odot \Delta^2 + \mF'_3 \odot \Delta^3$

\STATE \textbf{return} $\mZ^{\text{TS}}$
\end{algorithmic}
\end{algorithm}

\subsection{Cycle-Consistent Learning Mechanism}
\label{sub:cycle_consistency}

{The Cycle-Consistent Learning (CCL) mechanism incorporates a dance-to-music model to enhance temporal consistency between the generated dance sequences and the source music. The motivation is that a high-quality music-driven dance should preserve the information in the source music that is visually reflected by motion, such as rhythmic accents, temporal structure, and motion-related musical semantics. If the generated dance loses these music-related cues, the auxiliary D2M model cannot reconstruct the corresponding music features well. Therefore, reconstructing music features from generated motion provides a direct cross-modal supervision signal, encouraging the dance to preserve music-related information and remain aligned with the input music. Notably, we reconstruct music features rather than raw audio, because music features provide a compact and motion-relevant target while avoiding acoustic details that are weakly reflected by body motion. To achieve this, this mechanism employs an auxiliary task that first generates dance motion from input music, then reconstructs the music features from the generated dance with the dance-to-music model.}
\subsubsection{Dance-to-Music Generation}
\label{dance-to-music}
{We propose a Dance-to-Music (D2M) model to reconstruct music from the predicted dance motion $\hat{\mX} = [\hat{\vx}^1,\hat{\vx}^2,\cdots,\hat{\vx}^{T}]$.
Specifically, the model consists of a Variational Autoencoder (VAE) and a Latent Diffusion Model (LDM). To obtain high-quality music latents $\hat{\mM} = [\hat{\vm}^1,\hat{\vm}^2,\cdots,\hat{\vm}^{T'}]$ and achieve high-fidelity reconstruction, we adopt a pretrained VAE, which is trained on 20,000 mel-spectrograms, each representing a 5-second music clip randomly sampled from a Spotify playlist. Subsequently, a latent diffusion model leverages an autoencoder $\epsilon_\theta^l$ to predict the noise, conditioned on the dance motion via a cross-attention module. Similar to the music-to-dance model, we use the MSE loss to optimize $\epsilon_\theta^l$, which can be computed as:}
\begin{equation}
    \mathcal{L}_{\text{d2m}} = \mathbb{E}_{\hat{\mM}, \hat{\mX}, \epsilon, t}\Big[ \Vert \epsilon - \epsilon_\theta^l(\hat{\mM}_t, t, \hat{\mX}) \Vert_{2}^{2}\Big].
\end{equation}

\subsubsection{Cycle-Consistent Learning}
\label{cycle}
{We leverage the D2M model to establish a cycle-consistent mapping. Specifically, we first pre-train the D2M model and freeze its parameters to improve M2D training efficiency and stability. During the M2D training phase, the predicted dance sequences $\hat{\mX}$ (or the denoised output $\epsilon_\theta^s(\mX_t, t, \mM)$) are fed into the frozen D2M model. Specifically, conditioned on the predicted dance sequence, the frozen D2M model performs one denoising step in the latent space to estimate the music latent representation $\mM_R$. Rather than performing a complete iterative reverse diffusion process, which would introduce substantial computational overhead and a long back-propagation path, we employ this single-step inference strategy during M2D training. We introduce a cycle consistency loss that minimizes the L2 distance between the reconstructed music latents $\mM_R$ and the original music latents $\mM_O$:}
\begin{equation}
    \mathcal{L}_\text{cycle} = \|\mM_R - \mM_O\|_2^2.
\end{equation}

{The gradients from the $\mathcal{L}_\text{cycle}$ are propagated back through the M2D model, effectively regularizing its training. This additional supervision further helps the M2D model generate dance motions that better align with the semantic structure, \eg, beats, of the music.}

\section{Experiments}
In this section, we present a comprehensive evaluation of BeatDance. We first describe the experimental settings, including the datasets, implementation details, and evaluation metrics used throughout our evaluation. We then compare our method against state-of-the-art approaches through both quantitative and qualitative analysis. Finally, we conduct ablation studies to analyze the contribution of each proposed component.
\subsection{Experimental Settings}
In this subsection, we describe the two benchmark datasets used for evaluation, specify the implementation details of our model, and introduce the evaluation metrics adopted to assess the quality of generated dances.

\subsubsection{Datasets}
AIST++~\cite{li2021ai} is a large-scale 3D dance motion dataset
reconstructed from multi-view dance videos.
It contains 1,408 sequences with a total duration of 5.2 hours, covering
10 genres of street dance performed by 30 subjects.
The genres span both old-school styles (Break, Pop, Lock, and Waack)
and new-school styles (Middle Hip-hop, LA-style Hip-hop, House, Krump,
Street Jazz, and Ballet Jazz).
Each genre contains both basic (85\%) and advanced (15\%) choreographies.
The accompanying music spans tempos from 80 to 135 BPM.
Per-frame annotations include SMPL pose parameters for 24 joints with
global translation, and 17 COCO-format joint locations in both 2D and 3D.
The dataset is carefully split to ensure no overlap of music or
choreography between training and test sets.
Following~\cite{tseng2023edge}, we adopt the same split and preprocessing,
adjusting all training samples to 5 seconds at 30 fps.

PopDanceSet~\cite{luo2024popdg} is a more recent dataset designed
to reflect the aesthetic preferences of contemporary audiences.
It is constructed by filtering dance videos from a popular online platform
using a popularity function.
The dataset totals 12,819 seconds across 19 music genres — from CPOP
and KPOP to house dance and rock — performed by 132 subjects.
Per-frame annotations include 24 SMPL pose parameters and 17 COCO-format
3D joint locations, extracted using a monocular pose estimation model.
Compared to AIST++, PopDanceSet spans a broader range of dance styles and
music genres, and features more intricate choreographic movements, posing
greater challenges for music-dance alignment.
For fair comparison, we follow its official data split and process all
samples to 5 seconds at 30 fps.

Together, these two datasets enable a comprehensive evaluation of
BeatDance: AIST++ serves as a well-established benchmark, while
PopDanceSet provides a more choreographically complex scenario.

\subsubsection{Implementation Details}
All experiments are conducted on 4 NVIDIA A6000 GPUs.
Our framework consists of two branches: the music-to-dance (M2D) generation branch and the auxiliary dance-to-music (D2M) branch used for cycle-consistent learning.
We describe the implementation details of each branch below.

\paragraph{Dance Representation}
The dance feature vector comprises 156 dimensions: a 3-dimensional root joint translation, 6-DoF rotations for 24 SMPL body joints (144 dimensions in total), and 9 binary contact indicators for the feet, hands, and neck.
For the Hierarchical Decoupled Attention module, the 24 body joints are partitioned into $N_s{=}8$ groups: 7 functionally relevant regions incorporating the head, spine, left arm, right arm, left leg, right leg, and root joint, along with an additional global full-body part.
Each group is processed by an independent linear projection to preserve part-specific motion characteristics.

\paragraph{Music Representation}
We use the pretrained Jukebox~\cite{dhariwal2020jukebox} model to extract music features, which yields 4,800-dimensional embeddings per frame.
These embeddings encode rich semantic and rhythmic information from the raw audio, providing a strong conditioning signal for the M2D generation branch.
All music clips are processed at 30 fps to match the dance frame rate, yielding 150 frames per 5-second clip.

\paragraph{Music-to-Dance Branch}
The M2D diffusion model adopts a cosine-based variance schedule with a cosine offset of 0.008 and sets the total denoising steps $T$ to 1000.
The denoising network is a 6-layer Transformer with latent dimensions of $8{\times}64$, where 8 corresponds to the number of body parts and 64 is the per-part feature dimensionality.
The local hierarchical window size $\tau$ is set to 30 frames, and the Taylor series expansion order for the temporal attention map is set to 3, both determined through ablation studies (Table \ref{tab:order} and Table \ref{tab:window_size}).
During training, we use the Adan~\cite{xie2024adan} optimizer with a learning rate of $2{\times}10^{-4}$, a weight decay of 0.02, and a batch size of 128.
During inference, we employ DDIM sampling with 50 steps, which reduces inference time by $20{\times}$ relative to standard DDPM sampling while maintaining comparable generation quality.

\paragraph{Dance-to-Music Branch}
The D2M branch synthesizes audio represented as mel-spectrograms with 256 Mel-frequency bands.
The VAE encoder and decoder each consist of four convolutional blocks for progressive downsampling and upsampling, respectively, compressing the mel-spectrogram into a compact latent representation.
The latent diffusion model adopts a U-Net architecture with a four-layer convolutional encoder and a symmetric four-layer decoder connected via skip connections, enabling multi-scale feature reuse.
For training, we employ the AdamW optimizer~\cite{loshchilov2017decoupled} with a learning rate of $1{\times}10^{-4}$, a weight decay of $10^{-6}$, and a batch size of 32.

\paragraph{Training Strategy}
The M2D model is the primary focus of this work and is trained with the support of the auxiliary D2M branch via a two-stage procedure.
In the first stage, the D2M branch is pre-trained independently to ensure it can provide reliable cycle-consistency supervision.
In the second stage, its parameters are frozen, and it is used exclusively to provide cycle-consistency supervision for M2D: dance sequences generated by M2D are passed through the frozen D2M branch to reconstruct the source music, and the resulting $\mathcal{L}_{\text{cycle}}$ is back-propagated to refine the M2D model.
Throughout training, only the M2D parameters are updated.

\subsubsection{Evaluation Metrics}
\label{subsubsec:evaluation_metrics}
We follow previous methods~\cite{tseng2023edge,luo2024popdg} to evaluate generated dances from three dimensions: physical plausibility (PFC and PBC), diversity (Div$_k$ and Div$_g$), and beat alignment with music (BAS).
These metrics characterize complementary aspects of visual dance quality: PFC and PBC reflect whether the motion appears physically plausible, Div$_k$ and Div$_g$ reflect the variety of generated movements, and BAS reflects whether visible motion accents follow the musical rhythm. Since no single automatic metric fully captures the perceptual quality of a dance, we interpret them jointly.

\paragraph{Physical Foot Contact (PFC)}
PFC evaluates the physical plausibility of lower-body dance motion.
It is computed as:
\begin{equation}
    \text{PFC} = \frac{1}{N \cdot \max_{1 \leq j \leq N} \|\overline{\va}_{\text{COM}}^j\|}
    \sum_{i=1}^{N} s^i,
\end{equation}
where $N$ is the number of frames and $s^i$ is the per-frame plausibility score:
\begin{equation}
    s^i = \|\overline{\va}_{\text{COM}}^i\| \cdot \|\vv_{\text{LF}}^i\| \cdot \|\vv_{\text{RF}}^i\|.
\end{equation}
Here, $\vv_{\text{LF}}^i$ and $\vv_{\text{RF}}^i$ denote the velocities of the left and right foot at frame $i$, and $\overline{\va}_{\text{COM}}^i$ is the center-of-mass (COM) acceleration with its vertical component clamped to be non-negative:
\begin{equation}
    \overline{\va}_{\text{COM}}^i = \begin{pmatrix}
        a_{\text{COM},x}^i \\
        a_{\text{COM},y}^i \\
        \max(a_{\text{COM},z}^i,\; 0)
    \end{pmatrix}.
\end{equation}
PFC is designed to expose foot-sliding artifacts. Intuitively, when the body accelerates, at least one foot should provide stable ground contact; a frame is therefore penalized when the COM accelerates while both feet move simultaneously. A lower PFC indicates fewer visually implausible foot-ground artifacts.

\paragraph{Physical Body Contact (PBC)}
PBC extends PFC to assess the physical plausibility of full-body dance motion.
An auxiliary function $f$ is first defined as:
\begin{equation}
    f(x, y, z) = \frac{\sum_{i=1}^{N}
        \|\overline{\va}_x^i\| \cdot \|\vv_y^i\| \cdot \|\vv_z^i\|}
        {\max_{1 \leq j \leq N} \|\overline{\va}_x^j\|},
\end{equation}
where $\overline{\va}_x$ and $\vv_y$ denote the acceleration and velocity of the joints specified in each argument of $f$.
PBC is then defined as:
\begin{equation}
    \begin{split}
    \text{PBC} = \frac{1}{N} \big[
        &-f(\text{root, lfoot, rfoot}) \\
        &+ f(\text{lchest, lhand, null}) \\
        &+ f(\text{rchest, rhand, null}) \\
        &+ f(\text{neck, head, null}) \big].
    \end{split}
\end{equation}
For example, in $f(\text{root, lfoot, rfoot})$, $\overline{\va}_x$ is the acceleration of the root joint, and $\vv_y$, $\vv_z$ are the velocities of the left and right foot, respectively.
PBC extends the foot-contact assessment to the full body by combining root-foot, chest-hand, and neck-head motion relationships. It reflects whether the generated whole-body movements are physically coordinated. Intuitively, a dance with a realistic PBC exhibits fewer foot-sliding artifacts and more natural movements in which the hands and head move coherently with the torso. Following POPDG~\cite{luo2024popdg}, values closer to the ground-truth PBC are considered better.

\paragraph{Diversity (Div$_k$ and Div$_g$)}
We evaluate diversity from two complementary perspectives: kinematic (Div$_k$) and geometric (Div$_g$).
Each score is computed as the average pairwise Euclidean distance between the feature vectors of all generated sequences in the test set, calculated independently on kinematic and geometric features.
Div$_k$ measures variation in motion dynamics, such as differences in velocity and acceleration patterns, whereas Div$_g$ measures variation in geometric pose and choreographic configurations. Visually, they indicate whether a model generates a broad range of movement styles and body shapes instead of repeatedly producing similar sequences. However, abnormally jittery or distorted motions can also inflate pairwise distances. Therefore, following EDGE~\cite{tseng2023edge}, we regard values close to the ground-truth diversity as preferable to blindly maximizing either score.

\paragraph{Beat Alignment Score (BAS)}
BAS measures the rhythmic synchronization between the generated dance and the input music.
It is defined as:
\begin{equation}
    \text{BAS} = \frac{1}{N_p} \sum_{i=1}^{N_p}
    \exp\!\left(-\frac{\min_{b_j^m \in B^m} \|b_i^p - b_j^m\|^2}{2\sigma^2}\right),
\end{equation}
where $B^p = \{b_i^p\}$ and $B^m = \{b_j^m\}$ are the sets of kinematic and music beats, $N_p$ is the number of kinematic beats, and $\sigma$ is a normalization hyperparameter.
The kinematic beats are extracted as local minima of the kinetic velocity curve, and the music beats are detected using the librosa library.
For each kinematic beat, BAS assigns a larger contribution when a nearby music beat exists; consequently, a higher BAS indicates tighter temporal synchronization. In visual terms, high BAS corresponds to salient motion accents, such as brief pauses, direction changes, or transitions, occurring near musical beats.

\subsection{Comparison with State-of-the-Art Methods}

In this subsection, we compare BeatDance against competitive baselines on two benchmark datasets. We first present quantitative results across all evaluation metrics, followed by qualitative visualizations to illustrate the superiority of our method in generating expressive and beat-consistent dance motions.
For a fair comparison, all methods use the same experimental setting: 4,800-dimensional Jukebox music embeddings per frame, the same 156-dimensional motion representation, and 5-second music and motion sequences sampled at 30 fps. We evaluate all generated motions using the same protocol and metrics described in Section~\ref{subsubsec:evaluation_metrics}.

\begin{table*}[!t]
    \centering
    \caption{Comparison with the state-of-the-art methods on PopDanceSet and AIST++ test sets. Metrics are marked as: $\uparrow$ higher is better, $\downarrow$ lower is better, $\rightarrow$ closer to GT is better. \textbf{Bold} (\underline{underlined}) indicates the best (second-best) results. $^\dagger$ denotes diffusion-based methods.}
    \label{tab:cmp_combined}
    \small
    \begin{tabular*}{\textwidth}{@{\extracolsep{\fill}} llccccc @{}}
        \hline\hline
        \multirow{2}{*}{Dataset} & \multirow{2}{*}{Method} & \multicolumn{2}{c}{Motion Quality} & \multicolumn{2}{c}{Motion Diversity} & \multicolumn{1}{c}{Alignment} \\
        \cline{3-7}
         & & PFC $\downarrow$ & PBC $\rightarrow$ & Div$_k$ $\rightarrow$ & Div$_g$ $\rightarrow$ & BAS $\uparrow$ \\
        \hline\hline

        \multirow{5}{*}{PopDanceSet}
        & Ground Truth & 1.5824 & 8.7365 & 9.0219 & 7.2931 & 0.174 \\
        \cline{2-7}
        & Bailando~\cite{siyao2022bailando} & 3.9751 & 4.8863 & 5.1835 & 5.4342 & 0.230 \\
        & EDGE$^\dagger$~\cite{tseng2023edge} & 3.8366 & 4.0348 & \underline{6.1709} & 5.7568 & 0.224 \\
        & POPDG$^\dagger$~\cite{luo2024popdg} & \underline{1.8253} & 5.9492 & \textbf{7.1342} & \underline{5.8314} & 0.233 \\
        & DanceEditor$^\dagger$~\cite{zhang2025danceeditor} & 1.8972 & \underline{7.8429} & 5.6338 & 5.2459 & \underline{0.238} \\
        & \textbf{Ours}$^\dagger$ & \textbf{1.7332} & \textbf{7.9080} & 5.7236 & \textbf{5.8507} & \textbf{0.244} \\
        \hline

        \multirow{5}{*}{AIST++}
        & Ground Truth & 1.1806 & 8.0353 & 8.5190 & 8.9384 & 0.453 \\
        \cline{2-7}
        & Bailando~\cite{siyao2022bailando} & 1.0633 & \underline{6.3632} & 14.8835 & \textbf{6.6585} & 0.474 \\
        & EDGE$^\dagger$~\cite{tseng2023edge} & 1.0792 & 5.1172 & 2.5946 & 2.4477 & 0.466 \\
        & POPDG$^\dagger$~\cite{luo2024popdg} & \underline{0.9952} & 4.7050 & 2.4775 & 2.7832 & 0.468 \\
        & DanceEditor$^\dagger$~\cite{zhang2025danceeditor} & 1.6540 & 4.6520 & \underline{2.7542} & \underline{2.7951} & \underline{0.475} \\
        & \textbf{Ours}$^\dagger$ & \textbf{0.9877} & \textbf{7.1591} & \textbf{2.7965} & 2.7441 & \textbf{0.476} \\
        \hline\hline
    \end{tabular*}
\end{table*}

\begin{figure*}[t]
\centering
\includegraphics[width=\textwidth]{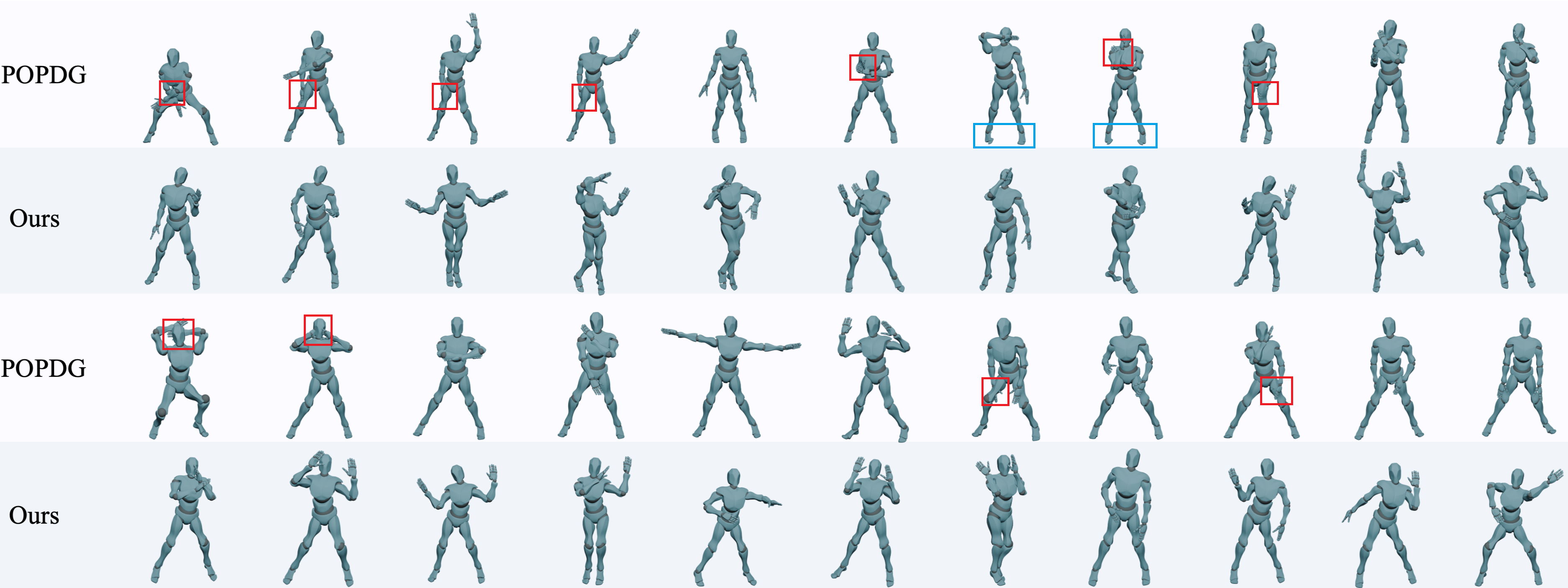}
\caption{Visual results on the PopDanceSet. We visually compare BeatDance with the state-of-the-art method POPDG~\cite{luo2024popdg}. Penetration artifacts (red) and Distorted motions (blue) are highlighted.}
\label{visualization}
\end{figure*}

\subsubsection{Quantitative Results}
\label{subsubsec:quant}

Table~\ref{tab:cmp_combined} presents quantitative evaluation results on the PopDanceSet and AIST++ datasets.
Our method achieves the best results on four of the five metrics on each dataset, including PFC, PBC, and BAS on both datasets. The remaining differences in diversity metrics are discussed below.

\paragraph{Motion quality}

BeatDance achieves state-of-the-art performance in both PFC and PBC on PopDanceSet and AIST++.
Physical plausibility is a foundational criterion in dance generation: motions with foot-sliding artifacts or implausible physical dynamics are immediately perceptible to viewers, rendering them unsuitable regardless of their beat alignment or diversity.
This improvement is driven by HDA and CCL in a complementary fashion. HDA models the fine-grained spatial-temporal structure of human motion. Its spatial branch learns region-specific physical dynamics through independent per-body-part projections. Its hierarchical temporal structure further ensures coherent full-body coordination across multiple scales. CCL additionally discourages physically implausible configurations by constraining the generated motion to remain semantically aligned with the source music. Together, they produce the state-of-the-art PFC and PBC scores across both benchmarks.

\paragraph{Beat alignment}

BeatDance also achieves state-of-the-art BAS on both PopDanceSet and AIST++.
As the defining objective of music-driven dance generation, beat alignment is indispensable: a dance that fails to respond to the rhythmic structure of the music defeats the fundamental purpose of the task, regardless of its diversity.
This advantage stems from the synergy between HDA and CCL. HDA equips the model with the capacity to learn fine-grained correspondences between musical rhythm and dance motion across multiple temporal scales, which is a prerequisite for precise beat synchronization. CCL then provides a direct optimization signal for this alignment. The music$\to$dance$\to$music round-trip can only close with low reconstruction error when the generated dance genuinely tracks the rhythmic structure of the source music. Together, the two components enforce beat-level synchronization.

\paragraph{Motion diversity}

BeatDance achieves the best geometric diversity (Div$_g$) on PopDanceSet and the best kinematic diversity (Div$_k$) on AIST++.
Diversity is \textit{at best an auxiliary indicator}, as pairwise distance can be inflated by physically implausible outlier poses.
A model generating degraded motions may therefore score artificially high on this metric, a phenomenon also noted in prior work~\cite{luo2024popdg,siyao2022bailando}.
For the two metrics where we do not rank first, the gaps are readily explained.
On PopDanceSet, POPDG achieves a higher Div$_k$ than ours (7.13 vs.\ 5.72 for BeatDance), but this comes at the cost of substantially lower motion quality (PBC 5.95 for POPDG vs.\ 7.91 for BeatDance) and beat alignment (BAS 0.233 for POPDG vs.\ 0.244 for BeatDance).
As shown in Figure~\ref{visualization}, POPDG's generated motions exhibit clear quality issues, including penetration artifacts and distorted poses.
Such implausible poses may partially account for its higher Div$_k$, as outlier motions tend to inflate pairwise distance measures.
On AIST++, our Div$_g$ (2.74) is competitive with the leading diffusion-based result (2.80).
The only method that clearly surpasses ours is Bailando, which achieves a Div$_g$ of 6.66, but this advantage does not generalize.
Bailando achieves the second-lowest Div$_g$ scores on PopDanceSet, suggesting that its high AIST++ diversity may reflect dataset-specific overfitting of its discrete choreographic memory rather than genuine expressive capability.

\begin{figure}[t]
\centering
\includegraphics[width=0.55\columnwidth]{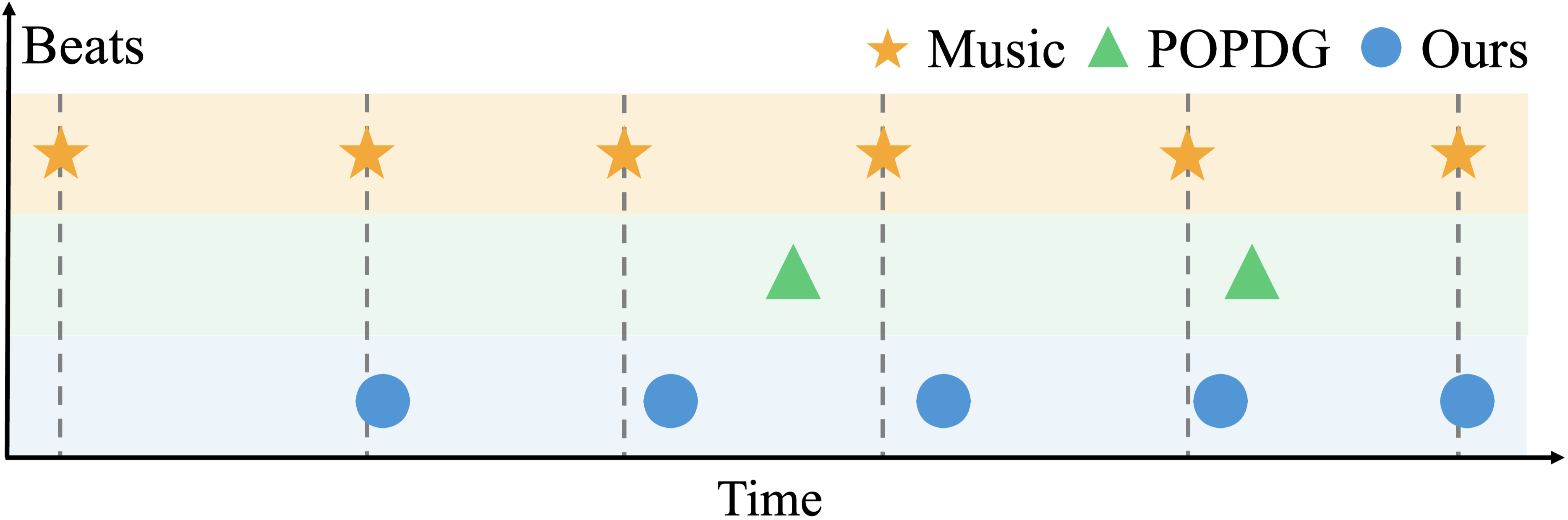}
\caption{Visual comparison of beat consistency.}
\label{beat}
\vspace{-0.5cm}
\end{figure}

\subsubsection{Qualitative Results}

Figure~\ref{visualization} shows the qualitative comparison on the PopDanceSet dataset.
Compared to POPDG~\cite{luo2024popdg}, our method generates more expressive dance motions. We highlight penetration artifacts and distorted motions using red and blue rectangles, respectively. Furthermore, we provide a beat synchronization analysis in Figure~\ref{beat}. Our model (blue circles) exhibits significantly higher rhythmic consistency than POPDG (green triangles). Specifically, BeatDance captures 5 of 6 musical beats, whereas POPDG only aligns with 2. These results further validate the effectiveness of HDA and CCL, which enhance the dance motion quality and the music-dance consistency, particularly in terms of beat-level alignment.
It is worth noting that Figure~\ref{beat} is provided only as an illustrative example for an intuitive understanding of beat alignment. The BAS results in Table~\ref{tab:cmp_combined} are computed over the complete test sets, where BeatDance achieves the best performance on both PopDanceSet and AIST++.
Regarding diversity, we observe that even though POPDG scores higher on Div$_k$, the qualitative results in Figure~\ref{visualization} reveal that a portion of its apparent kinematic variety arises from physically implausible configurations (\eg, penetration artifacts and distorted limb poses highlighted in red and blue), rather than genuine choreographic diversity. BeatDance, by contrast, maintains consistent limb topology across generated sequences while still producing varied whole-body postures adapted to the input music, suggesting that its lower Div$_k$ reflects tighter physical constraints rather than a lack of expressive range.

\subsubsection{User Study}
To complement the objective metrics, we conducted a pairwise user study with 20 participants using samples from the PopDanceSet test set. BeatDance was compared with POPDG and DanceEditor across multiple music clips, yielding 10 comparisons per participant and 200 pairwise judgments in total. Each pair consisted of a BeatDance result and a result from one competing method generated from the same music clip. For each pair, participants selected the dance with better overall quality while considering motion naturalness, expressiveness, and music-motion compatibility. Method identities were hidden, and both the comparison order and the left-right placement of the videos were randomized. As shown in Table~\ref{tab:user_study}, BeatDance was preferred over both competing methods, demonstrating its perceptual advantage in overall dance quality.

\begin{table}[H]
    \centering
    \caption{Pairwise user-study results on the PopDanceSet test set.}
    \label{tab:user_study}
    \small
    \begin{tabular}{lc}
        \hline\hline
        Comparison & BeatDance Win Rate \\
        \hline\hline
        BeatDance vs.\ POPDG & 91\% \\
        BeatDance vs.\ DanceEditor & 94\% \\
        \hline\hline
    \end{tabular}
\end{table}

\subsection{Ablation Studies}

To ensure experimental efficiency while maintaining a fair comparison, we adopt a scaled-down model configuration (half dimension and layers) for all ablation experiments, following the protocol in POPDG~\cite{luo2024popdg}. We empirically observed that the relative performance trends remain consistent between the scaled-down and full models.

\begin{table}[!t]     \centering
    \caption{Ablation study of different components. DA, HM, and CCL denote the Decoupled Attention, Hierarchical Mechanism, and Cycle Consistency Learning, respectively.}
    \label{tab:abl}
    \small
    \begin{tabular}{lccccc}
        \hline\hline
        Method & PFC & PBC & Div$_k$ & Div$_g$ & BAS \\
        \hline\hline
        w/o DA & 15.5267 & \underline{6.3755} & \textbf{8.9975} & 6.1423 & 0.218 \\
        w/o HM & 1.6046 & 5.7520 & \underline{6.9480} & \textbf{6.3608} & 0.240 \\
        w/o CCL & \underline{1.5594} & 5.8300 & 6.6467 & 5.5107 & \underline{0.252} \\
        Ours & \textbf{1.5390} & \textbf{7.7493} & 6.7770 & \underline{6.1801} & \textbf{0.273} \\
        \hline\hline
    \end{tabular}
\end{table}

\subsubsection{Investigating Each Component}

\paragraph{Effect of Decoupled Attention}

We replace DA with vanilla cross-attention to assess its contribution.
As shown in Table~\ref{tab:abl}, removing DA causes a dramatic drop in PFC
(1.54 $\to$ 15.53), PBC (7.75 $\to$ 6.38), and BAS (0.273 $\to$ 0.218).
In DA, each anatomical group is assigned an independent linear projection,
preserving part-specific dynamics and allowing each body region
to independently align with musical rhythm.
Replacing it with shared attention entangles the representations of all body
parts, disrupting both local physical constraints and fine-grained
music-motion correspondence.
Although \textit{w/o DA} achieves the highest Div$_k$ (8.9975), this is
consistent with the pattern observed in Section~\ref{subsubsec:quant}: physically
implausible motions tend to inflate pairwise kinematic distance.
The near-unchanged Div$_g$ (6.14 vs.\ 6.18) further supports this
interpretation, as geometric diversity is less sensitive to such artifacts.

\paragraph{Effect of Hierarchical Mechanism}

We remove the short-term branch of HM, retaining only long-term attention.
As shown in Table~\ref{tab:abl}, the primary degradations appear in PBC
(7.75 $\to$ 5.75) and BAS (0.273 $\to$ 0.240), while PFC changes only
marginally (1.54 $\to$ 1.60).
This pattern aligns with the design motivation of HM: musical beats are
sub-second local events that global attention, spread over the entire
sequence, lacks the temporal resolution to capture precisely.
The local windows (size $\tau{=}30$ frames, \ie, 1 second) provide
short-range receptive fields that improve beat-level alignment and
full-body coordination, as reflected in the BAS and PBC gains.
PFC, which primarily depends on part-specific spatial modeling already
provided by DA, is less sensitive to the removal of local temporal context.
The slight increases in Div$_k$ and Div$_g$ (6.78 $\to$ 6.95 and
6.18 $\to$ 6.36) suggest that removing local constraints allows the model
to explore a slightly broader motion distribution, at the cost of
physical plausibility and rhythmic consistency.

\paragraph{Effect of Cycle-Consistent Learning}

We remove CCL and train the M2D model without cycle-consistency supervision.
As shown in Table~\ref{tab:abl}, removing CCL leads to consistent degradation
across nearly all metrics: PBC drops from 7.75 to 5.83, BAS from 0.273 to
0.252, Div$_k$ from 6.78 to 6.65, and Div$_g$ from 6.18 to 5.51, with only
PFC remaining largely unchanged (1.54 $\to$ 1.56).
This broad pattern suggests that CCL fosters holistic cross-modal alignment
between music and full-body motion, especially beat alignment.

\begin{table}[tb]
\centering
\caption{Ablation study of the order of the Taylor series expansion.}
\label{tab:order}
\small
\begin{tabular}{cccccc}
\hline\hline
Order & PFC & PBC & Div$_k$ & Div$_g$ & BAS \\
\hline\hline
1 & \underline{1.2401} & 7.2673 & \underline{6.6585} & \underline{5.8027} & 0.229 \\
2 & 1.3925 & 6.9268 & 6.6322 & 5.4058 & 0.242 \\
3 & 1.5390 & \textbf{7.7493} & \textbf{6.7770} & \textbf{6.1801} & \textbf{0.273} \\
4 & 1.7447 & \underline{7.3459} & 6.5358 & 5.6094 & 0.255 \\
5 & \textbf{1.1339} & 6.4334 & 6.2685 & 5.1131 & \underline{0.256} \\
\hline\hline
\end{tabular}
\end{table}

\subsubsection{Effect of Taylor Series Expansion Order}
We investigate the effect of the Taylor series expansion order used to approximate the joint music-dance attention map.
As shown in Table~\ref{tab:order}, order 3 achieves the best performance across most metrics, including PBC (7.75), Div$_k$ (6.78), Div$_g$ (6.18), and BAS (0.273).
Lower orders (1 and 2) yield consistently lower BAS (0.229 and 0.242), suggesting that a low-order approximation lacks the capacity to capture the complex non-linear dependencies between music and dance, resulting in weaker beat alignment.
As the order increases beyond 3, performance degrades across most metrics:
PBC drops from 7.75 to 6.43 and BAS from 0.273 to 0.256 at order 5,
indicating that excessively high-order approximations introduce optimization difficulties that impair music-dance correspondence.
Although order 5 achieves the best PFC (1.13), this comes at the cost of significant degradation in all other metrics and therefore does not represent an overall improvement.
We thus adopt order 3 as the default setting.

\subsubsection{Effect of Local Window Size}

We analyze the effect of the local window size $\tau$ on model performance.
As shown in Table~\ref{tab:window_size}, a small window ($\tau{=}5$) achieves
the highest PBC but the worst PFC, as the limited temporal context is
insufficient to maintain global motion coherence.
Conversely, a large window ($\tau{=}50$) yields the best PFC but the worst
BAS (0.218), as the coarse temporal resolution makes it difficult to resolve
sub-second beat events.
Our chosen window size of 30 achieves the best BAS (0.273) and Div$_g$
(6.18) while maintaining competitive performance across other metrics,
achieving a favorable balance between local rhythmic precision and
long-range motion coherence.
Given that tempo varies across clips, $\tau{=}30$ should be understood
as an approximate temporal context rather than a fixed musical unit such as a
beat, bar, or phrase. This one-second context can cover local beat-level patterns
and short motion transitions while avoiding overly coarse temporal context that
weakens sensitivity to beat events.
We use $\tau{=}30$ by default.

\begin{table}[tb]
\centering
\caption{Ablation study of the local window size $\tau$.}
\label{tab:window_size}
\small
\begin{tabular}{cccccc}
\hline\hline
$\tau$ & PFC & PBC & Div$_k$ & Div$_g$ & BAS \\
\hline\hline
5  & 2.6125 & \textbf{8.5605} & \underline{7.5657} & 5.8857 & 0.232 \\
15 & 1.6489 & 6.9349 & 6.6073 & 5.5570 & \underline{0.249} \\
30 & \underline{1.5390} & \underline{7.7493} & 6.7770 & \textbf{6.1801} & \textbf{0.273} \\
50 & \textbf{1.4113} & 6.7600 & \textbf{8.7246} & \underline{6.0828} & 0.218 \\
75 & 1.8742 & 5.3648 & 7.2374 & 5.5517 & 0.234 \\
\hline\hline
\end{tabular}
\end{table}

\begin{table}[tb]
    \centering
    \caption{Ablation study of different generative architectures for the dance-to-music generation task.}
    \label{tab:music2dance}
    \small
    \begin{tabular}{ccccc}
        \hline\hline
        Method & BCS $\uparrow$ & BHS $\uparrow$ & FAD $\downarrow$ & BAS $\uparrow$ \\
        \hline\hline
        Ground Truth & 100 & 100 & 0 & 0.174 \\
        GAN & 86.5 & 52.8 & 7.63 & 0.191 \\
        VAE+GAN & 100.2 & 54.6 & 7.52 & 0.193 \\
        Diffusion & 110.8 & 62.1 & 7.60 & 0.193 \\
        VAE+Diffusion & \textbf{113.8} & \textbf{64.8} & \textbf{6.86} & \textbf{0.195} \\
        \hline\hline
    \end{tabular}\end{table}

\subsubsection{Effect of Generative Architecture}
To choose the most appropriate architecture for the dance-to-music branch, we compare two widely used architectures: Generative Adversarial Network (GAN)~\cite{goodfellow2020generative} and Diffusion models~\cite{ho2020denoising}. Each model is evaluated under two settings: operating on raw data and within a VAE~\cite{kingma2013auto} latent space. To evaluate rhythmic consistency and content similarity between the generated music and the ground truth, we adopt four metrics: BAS, BCS (Beats Coverage Score)~\cite{lee2019dancing}, BHS (Beats Hit Score)~\cite{lee2019dancing}, and FAD (Fréchet Audio Distance)~\cite{kilgour2018fad}. BCS and BHS evaluate rhythmic fidelity, where BCS measures the difference in total beat count between the synthesized and reference music, and BHS examines their temporal beat alignment. FAD is used to quantify content similarity and overall audio quality. As shown in Table~\ref{tab:music2dance}, the VAE + Diffusion configuration achieves the best performance across all metrics, indicating its superior capability in modeling both rhythm and audio content.

\section{Conclusion}
In this paper, we propose a diffusion-based framework for 3D music-driven dance generation, which comprises a Hierarchical Decoupled Attention (HDA) and a cycle-consistent learning (CCL) mechanism.
We employ HDA to capture both long-term and short-term dependencies between the source music and the generated dance by explicitly disentangling spatial and temporal modeling.
Additionally, CCL effectively enhances the music-dance consistency by an auxiliary music-to-dance-to-music task.
Consequently, our proposed approach outperforms recent competitive methods on two benchmark datasets, particularly in physical plausibility and beat alignment.
Extensive experiments also show its superior performance in maintaining beat consistency.
Despite these promising results, BeatDance still has several limitations.
First, the effectiveness of CCL depends on the reliability of the auxiliary D2M
model. Although the D2M model is independently pre-trained and the ablation
study validates the practical benefit of CCL, inaccurate D2M estimation under
challenging music patterns may introduce noisy or biased supervision. Second,
BeatDance may show weaker beat synchronization when the input contains unseen
music genres with highly complex rhythmic patterns. Future work will explore more
robust D2M supervision, explicit style control, and stronger generalization to
complex musical rhythms.

\section*{Acknowledgment}
This work was supported in part by the National Natural Science Foundation of China under Grant 62502447 and the Earth System Big Data Platform of the School of Earth Sciences, Zhejiang University.

\bibliographystyle{elsarticle-num}
\bibliography{ref}

\end{document}

%% file: math_commands.tex
\usepackage{amsmath,amsfonts,bm}

\def\eqref#1{equation~\ref{#1}}

\def\1{\bm{1}}

\def\va{{\bm{a}}}

\def\vc{{\bm{c}}}

\def\vm{{\bm{m}}}

\def\vt{{\bm{t}}}

\def\vv{{\bm{v}}}

\def\vx{{\bm{x}}}

\def\mA{{\bm{A}}}

\def\mF{{\bm{F}}}

\def\mK{{\bm{K}}}

\def\mM{{\bm{M}}}

\def\mT{{\bm{T}}}

\def\mV{{\bm{V}}}
\def\mW{{\bm{W}}}
\def\mX{{\bm{X}}}

\def\mZ{{\bm{Z}}}

\DeclareMathAlphabet{\mathsfit}{\encodingdefault}{\sfdefault}{m}{sl}
\SetMathAlphabet{\mathsfit}{bold}{\encodingdefault}{\sfdefault}{bx}{n}